\documentclass{article}

\usepackage[ruled,vlined]{algorithm2e}
\usepackage{arxiv}
\usepackage{graphicx}
\usepackage{subcaption}
\usepackage[utf8]{inputenc} 
\usepackage[T1]{fontenc}    
\usepackage{hyperref}       
\usepackage{url}            
\usepackage{booktabs}       
\usepackage{amsfonts}       
\usepackage{nicefrac}       
\usepackage{algorithm2e}    
\usepackage{microtype}      
\usepackage{lipsum}
\usepackage{graphicx}
\usepackage{amsmath} 
\usepackage{algorithm2e}
\usepackage{algpseudocode}

\graphicspath{ {./images/} }

\title{Quantification of Uncertainties in Probabilistic Deep Neural Network by Implementing Boosting of Variational Inference}

\author{{Pavia Bera}\\
School of Electrical Engineering\\
University of South Florida\\
Tampa, Florida 33620\\
Email: paviabera@usf.edu
\and
{Sanjukta Bhanja}\\
School of Electrical Engineering\\
University of South Florida\\
Tampa, Florida 33620\\
Email: bhanja@usf.edu
}

\begin{document}
\maketitle
\begin{abstract}
Modern neural network architectures have achieved remarkable accuracies but remain highly dependent on their training data, often lacking interpretability in their learned mappings. While effective on large datasets, they tend to overfit on smaller ones. Probabilistic neural networks, such as those utilizing variational inference, address this limitation by incorporating uncertainty estimation through weight distributions rather than point estimates. However, standard variational inference often relies on a single-density approximation, which can lead to poor posterior estimates and hinder model performance.

We propose \textbf{Boosted Bayesian Neural Networks (BBNN)}, a novel approach that enhances neural network weight distribution approximations using Boosting Variational Inference (BVI). By iteratively constructing a mixture of densities, BVI expands the approximating family, enabling a more expressive posterior that leads to improved generalization and uncertainty estimation. While this approach increases computational complexity, it significantly enhances accuracy—an essential tradeoff, particularly in high-stakes applications such as medical diagnostics, where false negatives can have severe consequences.

Our experimental results demonstrate that BBNN achieves \textbf{$\sim$5\%} higher accuracy compared to conventional neural networks while providing superior uncertainty quantification. This improvement highlights the effectiveness of leveraging a mixture-based variational family to better approximate the posterior distribution, ultimately advancing probabilistic deep learning.
 
\end{abstract}


\section{Introduction}
\label{intro}
The classical neural network was first proposed by McCullough and Walter Pitts in 1943 \cite{McCulloch1943}, marking a significant milestone in the development of artificial intelligence. These early models laid the foundation for modern neural networks, which have since evolved into deep learning architectures that achieve state-of-the-art performance across various domains \cite{Goodfellow2016, Bishop2006}. However, despite their success, deep neural networks (DNNs) remain heavily reliant on large amounts of labeled training data, leading to challenges in generalization, particularly in data-scarce environments \cite{Gal2016}.

One of the fundamental issues with conventional DNNs is their inability to quantify uncertainty in predictions. Standard neural networks employ deterministic weight estimates, making them susceptible to overfitting and poorly calibrated confidence estimates \cite{Srivastava2014, Kendall2017}. This limitation is particularly problematic in high-stakes applications such as medical diagnosis \cite{Smith2018}, financial modeling \cite{Hernandez2015}, and autonomous systems \cite{McAllister2017}, where uncertainty estimation is crucial for decision-making. Bayesian Neural Networks (BNNs) address this issue by treating weights as probability distributions rather than fixed values, thereby incorporating an inherent mechanism for uncertainty quantification \cite{Blundell2015}.

Variational inference (VI) has emerged as a scalable approach to approximating Bayesian posterior distributions in neural networks. However, traditional VI techniques often rely on simple variational families that fail to capture complex, multimodal posteriors \cite{Miller2017}. This shortcoming can lead to suboptimal generalization, as the approximated posterior lacks the flexibility to model uncertainty effectively. Furthermore, standard Bayesian methods such as Markov Chain Monte Carlo (MCMC) can be computationally expensive, limiting their applicability to large-scale deep learning models \cite{MacKay1992}.

To address these limitations, we propose \textbf{Boosted Bayesian Neural Networks (BBNNs)}, which leverage \textbf{Boosting Variational Inference (BVI)} to enhance weight distribution approximations. Unlike conventional variational methods that use a single variational density, BVI iteratively constructs a mixture of densities, enabling a more expressive and accurate posterior representation \cite{Hernandez2015}. This approach provides a flexible alternative to traditional inference techniques, allowing neural networks to capture a broader range of uncertainties while maintaining computational efficiency \cite{Foong2020}.

BBNNs not only provide superior uncertainty quantification but also demonstrate improved generalization in limited-data settings. Traditional deep learning models require vast amounts of labeled data to generalize effectively, whereas BBNNs, through their probabilistic framework, are inherently more robust to data scarcity \cite{Rasmussen2006}. This makes them particularly valuable in biomedical applications, fraud detection \cite{Bonawitz2019}, and adversarially robust learning systems \cite{Wilson2020}. Moreover, Bayesian learning methods have been shown to mitigate adversarial vulnerabilities by incorporating uncertainty awareness in decision-making \cite{Gal2016Dropout}.

Beyond uncertainty quantification, BBNNs offer several computational advantages. By integrating boosting techniques into variational inference, BBNNs strike a balance between model complexity and computational efficiency \cite{BayesByBackprop}. Unlike standard BNNs, which often require extensive MCMC sampling \cite{MacKay1992}, BVI enables a more scalable and computationally feasible approximation of the posterior. Recent advancements in Bayesian deep learning, such as Gaussian processes \cite{Rasmussen2006} and Bayesian dropout \cite{Gal2016Dropout}, further highlight the practical relevance of our proposed approach.

Another key advantage of BBNNs is their adaptability to dynamic environments. Traditional deep learning models often struggle with distribution shifts and non-stationary data. However, BBNNs dynamically update their uncertainty estimates, making them well-suited for real-time decision-making applications, such as robotics \cite{Kwiatkowski2019}, edge AI \cite{Lane2019}, and federated learning \cite{Bonawitz2019}. The ability to continuously refine posterior approximations ensures that BBNNs remain robust even in evolving data landscapes.

BBNNs also contribute to the broader field of Bayesian optimization. Bayesian optimization techniques, which rely on probabilistic models to guide the search for optimal hyperparameters, have been widely applied in deep learning \cite{Osborne2009, Snoek2012}. Our method enhances Bayesian optimization by incorporating a more accurate uncertainty model, improving sample efficiency and convergence rates. This makes BBNNs a powerful tool for tasks requiring adaptive learning, such as automated machine learning (AutoML) and self-improving AI systems.

Moreover, BBNNs can be integrated with Bayesian active learning strategies to improve data efficiency. Active learning, which involves selecting the most informative samples for labeling, benefits from accurate uncertainty estimates. By leveraging the uncertainty-aware properties of BBNNs, active learning models can significantly reduce the amount of labeled data required for training while maintaining high predictive performance \cite{Settles2009}. This is particularly useful in domains where data annotation is expensive, such as medical imaging and genomics.

Furthermore, Bayesian learning has gained traction in neuroscience and cognitive modeling. The brain is often hypothesized to function as a Bayesian inference engine, continuously updating beliefs based on incoming sensory data \cite{Doya2007}. By adopting a probabilistic learning framework, BBNNs align more closely with biological neural systems, making them valuable for modeling human decision-making and perception. Additionally, Bayesian methods have shown promise in reinforcement learning \cite{Ghavamzadeh2015}, where they enable more efficient exploration strategies and adaptive behavior in robotic systems.

Scaling Bayesian inference to deep learning remains an ongoing challenge due to the high-dimensional nature of modern neural networks. Techniques such as stochastic optimization and local reparameterization tricks \cite{Kingma2015} have improved the scalability of Bayesian methods, but computational efficiency remains a key concern. Future research in Bayesian deep learning will likely focus on developing more efficient inference techniques, potentially leveraging quantum computing and neuromorphic hardware to accelerate probabilistic computations \cite{Schuld2020}.

In this work, we present a comprehensive analysis of BBNNs, demonstrating their superiority in terms of predictive accuracy and uncertainty quantification. Our experimental results indicate that BBNNs achieve \textbf{$\sim$5\%} higher accuracy compared to conventional neural networks while maintaining significantly better-calibrated uncertainty estimates. This improvement underscores the effectiveness of leveraging a mixture-based variational family for posterior approximation, ultimately advancing probabilistic deep learning.

The rest of this paper is structured as follows. 
Section \ref{background} provides an overview of the background and related work, highlighting the foundational research that this study builds upon and extends. 
Section \ref{methodology} details the theoretical foundations of BVI and its integration into BBNNs. Section \ref{results} presents empirical evaluations across multiple benchmark datasets. Finally, Section \ref{conclusion} concludes with a summary of key contributions and insights.

Figure 1 illustrates the conceptual framework of BBNNs and their distinction from conventional neural networks.

\begin{figure}
  \includegraphics[width=\columnwidth]{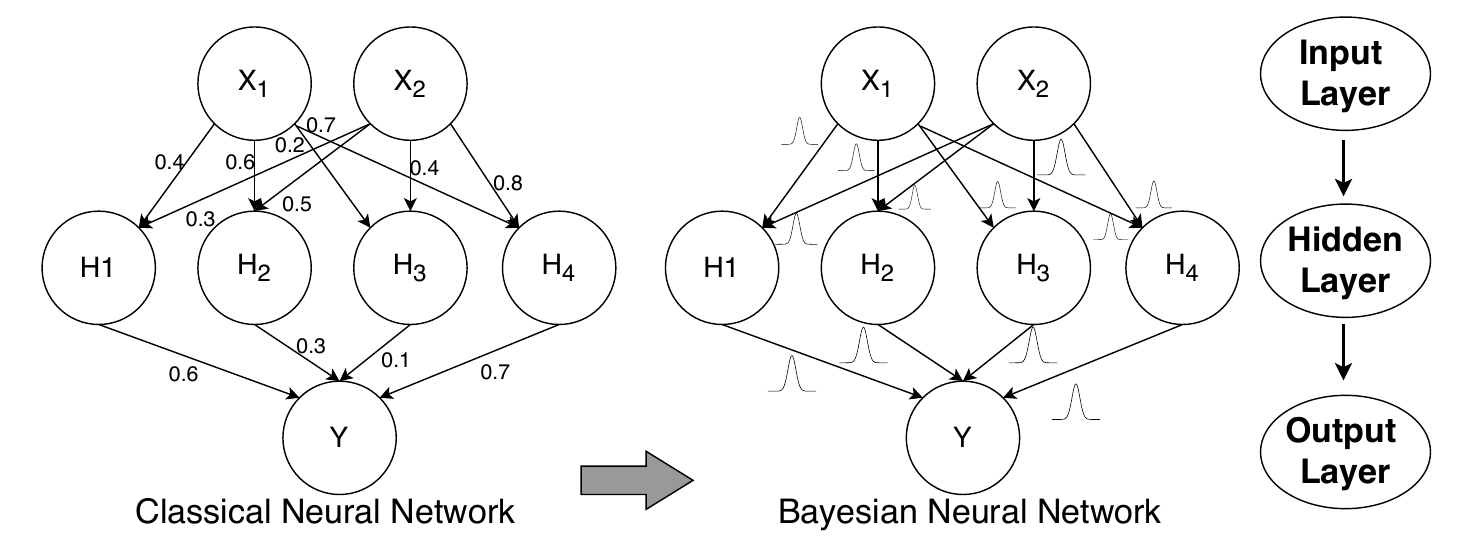}
  
\caption{Classical Neural Network with Point Weights and Neural Network with Distribution of Weights}
\label{fig:1}       
\end{figure}

\section{Background and Related Work}
\label{background}

The Bayesian approach to neural networks has gained significant attention in recent years due to its ability to model uncertainty in predictions while maintaining strong generalization performance. Traditional deep learning methods rely on point estimates for neural network weights, which often results in overfitting and poorly calibrated confidence estimates \cite{Goodfellow2016}. Bayesian neural networks (BNNs) overcome these challenges by treating network parameters as probability distributions, allowing the model to incorporate prior knowledge and effectively manage uncertainty in complex environments \cite{Bishop2006}.

A comprehensive review of Bayesian methods for neural networks is provided by Lampinen and Vehtari \cite{Lampinen2001}, where the authors discuss various approaches to Bayesian learning, emphasizing the role of prior knowledge in statistical modeling. They highlight how the Bayesian framework naturally incorporates model complexity through hierarchical priors and posterior distributions, which dynamically adjust based on data observations. Unlike classical error minimization methods, Bayesian inference provides a more principled approach to generalization by integrating uncertainty across different model complexities.

A key contribution of the Bayesian approach is its ability to propagate uncertainty, making it particularly useful in applications where robustness is critical. The review paper explores three case studies—regression, classification, and inverse problems—where Bayesian models outperform traditional neural networks by leveraging flexible priors and hierarchical structures \cite{Lampinen2001}. This aligns with the broader literature on Bayesian deep learning, where researchers have demonstrated that incorporating prior distributions over network weights improves model interpretability and decision-making reliability \cite{Neal1996, Blundell2015}.

One of the central challenges in Bayesian neural networks is computational efficiency. While traditional Markov Chain Monte Carlo (MCMC) methods provide an exact inference framework, they are often computationally intractable for large-scale neural networks. To address this, variational inference (VI) has emerged as a scalable alternative \cite{Kingma2015}. VI approximates the posterior distribution by optimizing a family of tractable distributions, enabling faster convergence while retaining the benefits of Bayesian modeling. Lampinen and Vehtari \cite{Lampinen2001} discuss the trade-offs between MCMC and VI, emphasizing the need for more efficient inference techniques to scale Bayesian methods to deep learning applications.

Recent advancements in boosting techniques have further improved the expressiveness of variational approximations. Boosting Variational Inference (BVI) iteratively refines the posterior approximation by constructing a mixture of densities, thereby enhancing flexibility and reducing approximation error \cite{Miller2017}. The integration of boosting methods into Bayesian neural networks has led to significant improvements in uncertainty estimation and generalization performance \cite{Hernandez2015}. These approaches align with the insights presented in the review paper, where hierarchical Bayesian models are shown to effectively manage model complexity and prevent overfitting.

Beyond the core theoretical developments, Bayesian neural networks have been applied to a wide range of real-world problems. Lampinen and Vehtari \cite{Lampinen2001} illustrate how Bayesian techniques improve model reliability in applications such as tomographic image reconstruction, quality assessment in manufacturing, and automated classification in forestry. These applications highlight the practical advantages of Bayesian modeling in domains where data is limited, noise levels are high, and interpretability is crucial.

Our work builds upon this foundational research by integrating \textbf{Boosted Bayesian Neural Networks (BBNNs)} with \textbf{Boosting Variational Inference (BVI)} to enhance the approximation of posterior distributions. By leveraging mixture-based variational families, BBNNs offer greater flexibility in weight uncertainty modeling, leading to superior predictive performance and robustness. The review paper serves as a valuable reference in understanding the evolution of Bayesian methods in neural networks, reinforcing the importance of hierarchical priors, posterior regularization, and scalable inference techniques.

\subsection{Probabilistic Backpropagation for Bayesian Neural Networks}

Hernández-Lobato and Adams proposed \textit{Probabilistic Backpropagation (PBP)} as an alternative to traditional Bayesian inference methods by propagating probability distributions rather than point estimates through the network \cite{Hernandez2015}. PBP reduces the variance of Monte Carlo approximations and improves scalability while maintaining high accuracy in posterior uncertainty estimation.

The key idea behind PBP is to approximate the posterior distribution of the weights, \( p(W | D) \), where \( D = \{(x_i, y_i)\}_{i=1}^{N} \) represents the training dataset. Instead of maintaining point estimates for the weights, PBP models them as Gaussian distributions with mean \( \mu_w \) and variance \( \sigma_w^2 \). The forward pass in PBP involves propagating these distributions through the network:

\begin{equation}
q(W) = \prod_{i=1}^{L} \mathcal{N}(W_i | \mu_{w_i}, \sigma_{w_i}^2)
\end{equation}

where \( L \) represents the number of layers in the network.

During training, PBP updates the variational parameters \( \mu_w \) and \( \sigma_w^2 \) using moment matching, an approach that minimizes the Kullback-Leibler (KL) divergence between the true posterior and the approximate posterior:

\begin{equation}
\text{KL}( q(W) || p(W | D) ) = \int q(W) \log \frac{q(W)}{p(W | D)} dW
\end{equation}

The update rules for the variational parameters are derived by minimizing this KL divergence while maximizing the likelihood of the observed data. The predictive distribution for a new input \( x_* \) is obtained by integrating over the posterior:

\begin{equation}
p(y_* | x_*, D) = \int p(y_* | x_*, W) q(W) dW
\end{equation}

This integral is typically approximated using a Gaussian distribution, leading to closed-form updates for the mean and variance of the predictive distribution.

One advantage of PBP over traditional Monte Carlo methods is its ability to approximate posterior distributions without requiring extensive sampling, thus significantly reducing computational overhead. Additionally, PBP can be efficiently implemented in deep networks by leveraging factorized Gaussian approximations, which allow for tractable backpropagation updates.

Experimental results demonstrate that PBP achieves state-of-the-art performance on various regression and classification tasks, offering both improved scalability and robust uncertainty estimation. The ability to model uncertainty effectively makes PBP particularly useful in safety-critical applications such as medical diagnosis and autonomous systems.

\subsection{Uncertainty Estimation via Softplus Normalization}

Shridhar et al. introduced \textit{Softplus normalization} as a means to estimate aleatoric and epistemic uncertainties in CNNs, improving model calibration without requiring additional softmax layers \cite{Shridhar2018}. Their method ensures a more coherent integration of uncertainty estimation in classification tasks.

Uncertainty in neural networks can be broadly classified into \textit{aleatoric} (data-dependent) and \textit{epistemic} (model-dependent) uncertainties. Aleatoric uncertainty arises from inherent noise in the data and is typically modeled using the variance of predictive distributions. Epistemic uncertainty, on the other hand, results from a lack of knowledge about the optimal model parameters and can be reduced with more data.

The Softplus function, defined as:

\begin{equation}
\text{Softplus}(x) = \log(1 + e^x),
\end{equation}

is used in place of standard activation functions to ensure positive and stable variance estimates. In Bayesian convolutional neural networks (BayesCNNs), the predictive distribution is modeled as a Gaussian with mean \( \mu(x) \) and variance \( \sigma^2(x) \), where the variance is estimated using Softplus normalization:

\begin{equation}
\sigma^2(x) = \text{Softplus}(f_{\sigma}(x)) = \log(1 + e^{f_{\sigma}(x)})
\end{equation}

where \( f_{\sigma}(x) \) is the output of the variance-estimating neural network. This formulation ensures that variance estimates remain positive and stable throughout training, preventing numerical instabilities.

The total predictive uncertainty in a Bayesian neural network can be decomposed as:

\begin{equation}
\text{Var}(y | x, D) = \underbrace{\mathbb{E}_{q(W)}[\sigma^2(x)]}_{\text{Aleatoric Uncertainty}} + \underbrace{\text{Var}_{q(W)}[\mathbb{E}[y | x, W]]}_{\text{Epistemic Uncertainty}}
\end{equation}

where \( q(W) \) represents the variational posterior over the network weights.

By incorporating Softplus normalization, the model can efficiently estimate both uncertainty types in a single forward pass without the need for additional sampling-based methods. This approach enhances the reliability of Bayesian CNNs in tasks requiring robust confidence estimation, such as medical image analysis and autonomous driving.

\subsection{Boosting Variational Inference for Expressive Posterior Approximations}

Miller et al. proposed \textit{Boosting Variational Inference (BVI)}, which iteratively refines the variational posterior by constructing a mixture of densities \cite{Miller2017}. This method enhances the expressiveness of variational approximations and reduces approximation error, leading to improved generalization in Bayesian neural networks.

Traditional variational inference (VI) methods approximate the posterior distribution \( p(W | D) \) with a single density \( q(W) \), which can lead to poor posterior estimates when the true distribution is complex or multimodal. BVI addresses this limitation by constructing a mixture of variational densities \( q_t(W) \), iteratively refining the approximation through boosting steps:

\begin{equation}
q_{t+1}(W) = (1 - \lambda_t) q_t(W) + \lambda_t q_{	ext{new}}(W),
\end{equation}

where \( q_{	ext{new}}(W) \) is the new variational component added at iteration \( t \), and \( \lambda_t \) controls the contribution of the new component to the overall approximation.

The objective function in BVI is based on minimizing the Kullback-Leibler (KL) divergence between the true posterior and the mixture approximation:

\begin{equation}
\text{KL}(q_{t+1}(W) || p(W | D)) = \mathbb{E}_{q_{t+1}(W)} \left[ \log \frac{q_{t+1}(W)}{p(W | D)} \right].
\end{equation}

Each new variational component \( q_{	ext{new}}(W) \) is selected to minimize the residual approximation error from previous steps, allowing BVI to iteratively refine its posterior estimation in a more flexible manner compared to standard VI.

One key advantage of BVI is its ability to capture multimodal posteriors, which is essential for Bayesian neural networks trained on complex datasets. Unlike traditional VI, which assumes a unimodal Gaussian approximation, BVI enables a more expressive representation of uncertainty, leading to improved predictive performance.

BVI has been successfully applied to Bayesian neural networks for both regression and classification tasks. In regression problems, the improved posterior estimation allows for more accurate uncertainty quantification, reducing overconfidence in out-of-distribution predictions. In classification tasks, BVI enhances model robustness by improving the separation between uncertain and confident predictions, which is crucial in applications such as medical diagnosis and autonomous systems.

Additionally, BVI provides a scalable alternative to Monte Carlo sampling methods, reducing the computational cost associated with traditional Bayesian inference. By leveraging a mixture of densities, BVI achieves a balance between flexibility and efficiency, making it an attractive choice for modern deep learning applications that require uncertainty-aware predictions.

Overall, BVI represents a significant advancement in variational inference for Bayesian neural networks, enabling more expressive posterior approximations and improved generalization. Its integration with Bayesian deep learning frameworks opens new possibilities for robust uncertainty estimation in high-dimensional models.

\section{Method}
\label{methodology}
Bayesian Neural Networks are capable of overcoming a substantial number of issues faced by classical or frequentist neural networks by introducing probabilistic training of weights. The probabilistic training makes use of Bayesian statistics to compute probability densities of the weights. This probability density is dependent on the posterior density. However, probabilistic models face a bottleneck when calculating the posterior. Bayes' theorem, stated in Equation (1), consists of the denominator \(p(b)\), which is intractable.

\begin{equation}
\begin{aligned}
p(a|b) = \frac{p(b|a) \cdot p(a)}{p(b)}
\end{aligned}
\end{equation}

\noindent where \( p(b) \) can be expressed as:

\begin{equation}
p(b)= \int_{a} p(b|a)p(a) da
\end{equation}

which is called the model evidence. The integral becomes intractable due to the high dimensionality of the parameter space and the non-linear functions of the neural network.

To approximate the intractable posterior \( p(a|b) \), different approaches have been proposed. \textit{Sampling-based methods}, such as Markov Chain Monte Carlo (MCMC), are a classical choice but suffer from high computational costs, making them impractical for deep networks. \textit{Optimization-based methods}, such as Variational Inference (VI), offer a more scalable alternative by transforming posterior approximation into an optimization problem.

\textit{Variational Inference (VI)} \cite{wainwright2008graphical, jordan1999introduction} has become the dominant approach for Bayesian deep learning due to its computational efficiency. Instead of directly computing the posterior, VI approximates it using a simpler distribution \( q(W) \), selected to minimize the \textit{Kullback-Leibler (KL) divergence} from the true posterior:

\begin{equation}
\text{KL}(q(W) || p(W|D)) = \mathbb{E}_{q(W)} \left[ \log \frac{q(W)}{p(W | D)} \right].
\end{equation}

Despite its advantages, \textit{VI suffers from a major limitation}: the approximating distribution \( q(W) \) is often \textit{too simple}, failing to capture complex or multimodal posteriors. This results in \textit{poor uncertainty estimation} and \textit{reduced predictive accuracy}.

To address this limitation, we introduce \textit{Boosting Variational Inference (BVI)}. Instead of using a single simple distribution to approximate the posterior, BVI \textit{iteratively refines the approximation} by constructing a mixture of densities:

\begin{equation}
q_{t+1}(W) = (1 - \lambda_t) q_t(W) + \lambda_t q_{\text{new}}(W)
\end{equation}

where \( q_{\text{new}}(W) \) is the newly added variational component, and \( \lambda_t \) determines its contribution to the overall mixture. This iterative refinement enables BVI to \textit{capture complex multimodal posteriors}, significantly improving approximation accuracy.

BVI is particularly well-suited for training Bayesian neural networks. The objective of training is to maximize the \textit{Evidence Lower Bound (ELBO)}:

\begin{equation}
\mathcal{L}(q) = \mathbb{E}_{q(W)}[\log p(D | W)] - \text{KL}(q(W) || p(W))
\end{equation}

where the first term represents the likelihood of the observed data under the posterior, and the second term enforces prior regularization via KL divergence. Unlike standard VI, which optimizes a single variational distribution, BVI \textit{iteratively improves} \( q(W) \) by introducing new variational components, making it \textit{more expressive and flexible}.

To integrate BVI into weight training, we modify standard \textit{backpropagation} by introducing boosting iterations. During each iteration, the gradient updates are modified to incorporate the newly introduced variational components:

\begin{equation}
\nabla_{\theta} \mathcal{L}(q) = \mathbb{E}_{q(W)} \left[ \nabla_{\theta} \log p(D | W) \right] - \nabla_{\theta} \text{KL}(q(W) || p(W))
\end{equation}

where \( \theta \) denotes the variational parameters. This formulation ensures that newly added components in BVI \textit{contribute towards refining the posterior approximation}, making it both accurate and computationally feasible.

In the following sections:
\begin{itemize}
    \item \textbf{Section 3.1} discusses \textit{Stochastic Variational Inference (SVI)} and its role in handling large datasets.
    \item \textbf{Section 3.2} introduces \textit{Boosting Variational Inference (BVI)} and details its implementation.
    \item \textbf{Section 3.3} explains \textit{how BVI is integrated into backpropagation} for efficient neural network training.
\end{itemize}

\subsection{Stochastic Variational Inference}
\label{sec:1}

The objective of \textit{Variational Inference (VI)} is to approximate the true posterior density from a family of target densities by minimizing the \textit{Kullback-Leibler (KL)} divergence \cite{kullback1951information}. The KL divergence quantifies the difference between the true posterior distribution and the variational approximation (Eq. 2). Lower KL divergence values indicate a better approximation:

\begin{equation}
q_{\lambda}(a) = \underset{q(a) \in Q}{\mathrm{arg \: min }} \:  KL[q(a)\:||\:p(b|a)]
\end{equation}

where \( b \) is the observed data, \( a \) is the set of latent variables, and the target distribution \( q(a) \) is parameterized by \( \lambda \). The goal is to bring \( q(a) \) as close as possible to the true posterior \( p(b|a) \).

By expanding the conditional in (Eq. 2), we can express the expectation in terms of the \textit{Evidence Lower Bound (ELBO)}:

\begin{align}
\mathbb{E}_{q_{\lambda}(a)} [\log p(a)] + \mathbb{E}_{q_{\lambda}(a)} [\log p(b|a)] - \mathbb{E}_{q_{\lambda}(a)} [\log q(a)] \\
= \mathbb{E}[\log p(b|a)] - KL [q(a)||p(a)]
\end{align}

Since KL divergence cannot be directly computed, ELBO is maximized (Eq. 5) as an alternative:

\begin{equation}
\mathcal{L}(\lambda) = \mathbb{E}_{q_{\lambda}(a)} [\log p(b,a) - \log q(a)]
\end{equation}

For models with a simple variational family, closed-form solutions exist \cite{ghahramani2001propagation}. However, for more complex models, closed-form solutions are not available, requiring iterative variational algorithms \cite{blei2017variational}. The expectation in (Eq. 5) can become computationally prohibitive when the variational family is complex.

To address this, researchers have employed model-specific algorithms \cite{jaakkola1997variational, blei2007correlated, braun2010variational} or general-purpose algorithms with model-specific computations \cite{knowles2011non, wang2013variational, paisley2012variational}. However, setting up these algorithms on a case-by-case basis is cumbersome.

To overcome this computational bottleneck, we employ \textit{Stochastic Variational Inference (SVI)}, which maximizes ELBO for general models using stochastic optimization. In SVI, noisy but unbiased gradients are sampled from the variational approximation to update the free parameter space \( \lambda \). The expectation with respect to the variational approximation defines the derivative of the objective function \cite{ranganath2014black}:

\begin{equation}
\Delta_{\lambda} \mathcal{L} = \mathbb{E}_{q} \left[\Delta_{\lambda} \log q(a|\lambda) \left(\log p (b,a)-\log q(a|\lambda)\right)\right]
\end{equation}

Since Eq. (7) directly links the score function and sampling to the variational distribution, SVI allows a black-box approach for posterior approximation. This reduces the need for model-specific parameter tuning and simplifies implementation across applications.

However, Eq. (7) can produce high-variance ELBO gradient estimates, slowing convergence due to small optimization steps. To counteract this, we introduce a \textit{control variate} based on the score function \( h \) of \( \Delta_{\lambda} \log q(a) \), which has zero expectation \cite{casella1996rao, paisley2012variational, ross2002Simulation}. Assuming the function whose expectation is computed is \( f_i \) and the control variate is \( h_i \), Eq. (8) refines the ELBO gradient estimate:

\begin{align}
   \begin{split}
f_i(a) = \Delta_{\lambda_i} \log q(a|\lambda_i) \left[\log (b,a) - \log q(a|\lambda_i)\right] \\
h_i(a) = \Delta_{\lambda_i} \log q(a|\lambda_i)
\end{split}
\end{align}

For \( d \) dimensions, the optimal scaling of the control variate is given by:

\begin{equation}
\hat{a_i^*} = \frac{\sum_{d=1} ^{n_i} \hat{\text{Cov}} (f_i^d,h_i^d)}{\sum_{d=1}^{n_i} \hat{\text{Var}}(h_i^d)}
\end{equation}

Substituting this into Eq. (7) produces a Monte Carlo estimate for the ELBO gradient:

\begin{equation}
\hat{\Delta}_{\lambda_i} \mathcal{L} = \frac{1}{S} \sum_{s=1}^{S} \left[\Delta_{\lambda} \log q(a_s|\lambda_i) \left(\log p_i (b,a_s)-\log q_i(a_s|\lambda_i) - \hat{a}_i^*\right)\right],
\end{equation}

where \( a_s \sim q_i(a|\lambda) \). Introducing \( \hat{a}_i^* \) requires computing only \( f_i \) and \( h_i \), which significantly reduces variance in gradient estimates without modifying the core inference algorithm.

Algorithm 1 summarizes the black-box variational inference process. The objective is to maximize ELBO \( \mathcal{L}(\lambda) \), using the score function estimation and Monte Carlo sampling.

\begin{algorithm}
\SetAlgoLined
\KwData{Observed data $b$, latent variable $a$}
\KwIn{Variational family $q$, joint distribution $p(a, b)$}
\KwOut{Optimized variational posterior $q(a)$}

\textbf{Objective}: Maximize the Evidence Lower Bound (ELBO) $\mathcal{L}(\lambda)$

\textbf{Initialize}: $\lambda_{1:n}$ (Random initialization), $t = 1$

\While {Change in $\lambda$ $<$ threshold}{

    Draw $S$ samples from $q(a|\lambda)$
    
    \For{$s = 1$ to $S$}{
    
        Sample $a_s \sim q(a|\lambda)$
    
    }
    
    Compute gradient estimate:
    
    \[
    \Delta \lambda = \frac{1}{S} \sum_{s=1}^{S} \left[ \nabla_{\lambda} \log q(a_s|\lambda) \left(\log p (b, a_s)-\log q(a_s|\lambda)\right) \right]
    \]
    
    Update $\lambda$ using learning rate $\rho_t$:
    
    \[
    \lambda \leftarrow \lambda + \rho_t \Delta \lambda
    \]
    
    $t \leftarrow t + 1$
    
}
\caption{Stochastic Variational Inference}

\end{algorithm}

This approach makes Variational Inference computationally efficient while maintaining flexibility across different models. By incorporating adaptive learning rates \cite{duchi2011adaptive}, the optimization process ensures robustness across different scales.

\subsection{Boosting of Variational Inference}
\label{sec:2}

\begin{figure}
  \includegraphics[width=\columnwidth]{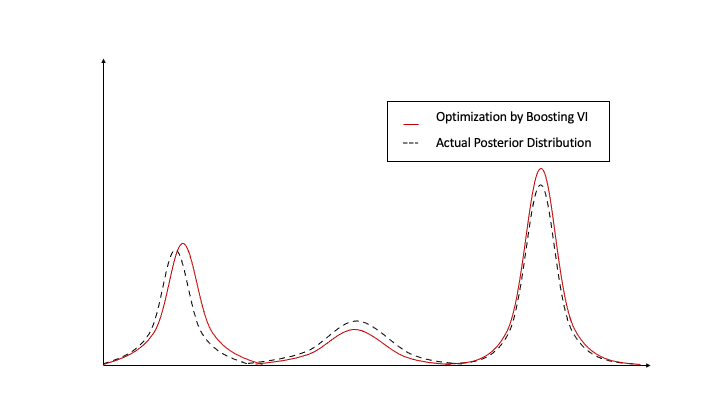}
  \caption{Optimization of the true 1-Dimensional posterior after adding components in Boosting Variational Inference.}
  \label{fig:3}
\end{figure}

One of the most significant limitations of Variational Inference (VI) is the constrained family of distributions from which the closest posterior distribution is optimized. Regardless of how extensively VI is run, it is never possible to fully capture the exact posterior distribution. \textit{Boosting Variational Inference (BVI)} \cite{guo2016boosting, miller2017variational} addresses this limitation by iteratively refining the approximate posterior distribution, incorporating additional components from a more flexible approximating family, as demonstrated in Figure~\ref{fig:3}. Unlike standard VI, which is restricted by its initial approximation choice, BVI progressively improves accuracy by introducing new components, making it particularly advantageous for capturing multi-modality in posterior distributions.

BVI offers an adaptive trade-off between computation time and accuracy. The more iterations that are expended on refinement, the more accurate the posterior approximation becomes. This is particularly beneficial when dealing with diverse datasets where standard VI methods struggle with complex posterior landscapes. Furthermore, traditional VI approximation methods \cite{bishop2006pattern, wang2005inadequacy, turner+sahani:2011a, rue2009approximate, mackay2003information} tend to miscalculate covariance in the posterior distribution, leading to suboptimal uncertainty estimation. Methods introduced by \cite{giordano2015linear, kucukelbir2015automatic} improved covariance estimation but still failed to incorporate multimodality, a key feature captured by BVI.

BVI is inspired by boosting techniques in machine learning, where optimization starts with a single simple component and progressively refines the model by adding new components from a family of distributions. This progressive learning strategy allows for more flexible approximations, particularly in cases where posterior distributions exhibit non-standard shapes. Unlike standard VI, which typically assumes a unimodal Gaussian posterior, BVI allows for the construction of more expressive approximations:

\begin{equation}
q_{t+1}(W) = (1 - \lambda_t) q_t(W) + \lambda_t q_{\text{new}}(W),
\end{equation}

where \( q_{\text{new}}(W) \) represents a newly introduced component, and \( \lambda_t \) controls its weighting in the mixture. This iterative refinement ensures that the posterior distribution is not constrained by a single approximating function but instead grows in complexity as needed.

The ELBO for BVI remains similar to standard VI but is adapted to accommodate the evolving mixture model:

\begin{equation}
\mathcal{L}(q) = \mathbb{E}_{q_{t+1}(W)}[\log p(D | W)] - \text{KL}(q_{t+1}(W) || p(W)).
\end{equation}

The integration of BVI into Bayesian Neural Networks provides several advantages:
\begin{itemize}
    \item \textbf{Capturing Multi-modality:} Unlike standard VI, which struggles with multi-modal distributions, BVI iteratively builds an approximation capable of representing multiple modes in the posterior.
    \item \textbf{Scalability and Flexibility:} By refining the posterior distribution progressively, BVI achieves a balance between computational feasibility and approximation accuracy.
    \item \textbf{Better Uncertainty Estimation:} Since BVI improves covariance structure estimation, it provides more reliable uncertainty quantification compared to traditional VI approaches.
\end{itemize}

\begin{algorithm}[H]
\SetAlgoLined
\KwData{Observed data $b$}
\KwIn{Initial variational approximation $q_0(W)$, prior $p(W)$, step size $\lambda_t$}
\KwOut{Boosted variational posterior $q_T(W)$}
\textbf{Objective}: Iteratively refine the variational approximation by adding components.

\textbf{Initialize}: Set $q_0(W)$ to an initial variational family, $t = 0$

\While {ELBO improvement $>$ threshold}{
    Compute residual error: \\
    $r_t(W) = p(W | b) - q_t(W)$ \\
    Fit a new variational component to residual: \\
    $q_{\text{new}}(W) \approx \underset{q}{\arg\min} \: \text{KL}(q(W) || r_t(W))$ \\
    Update the boosted posterior: \\
    $q_{t+1}(W) = (1 - \lambda_t) q_t(W) + \lambda_t q_{\text{new}}(W)$ \\
    $t \leftarrow t + 1$
}
\caption{Boosting Variational Inference}
\end{algorithm}

By incorporating BVI, Bayesian deep learning models can achieve a richer representation of posterior distributions, leading to improved performance in probabilistic inference tasks. The next section discusses how BVI is integrated into weight training, leveraging the reparameterization trick and Bayes By Backpropagation for efficient posterior refinement.

\subsection{Weight Training via the Reparameterization Trick and Bayes By Backpropagation}
\label{sec:3}

A key challenge in training Bayesian Neural Networks (BNNs) using Variational Inference (VI) is computing expectations over the variational posterior distribution. 
Given a dataset \( D = \{x_i, y_i\}_{i=1}^{N} \), where \( x_i \) represents input features and \( y_i \) denotes corresponding labels, our goal is to approximate the posterior distribution over network weights.
Directly sampling from the distribution introduces high variance in gradient estimates, making optimization unstable. To address this, the \textit{reparameterization trick} is used to enable low-variance gradient estimates and facilitate efficient backpropagation.

In traditional VI, the approximate posterior over weights \( W \) is typically assumed to follow a Gaussian distribution:

\begin{equation}
q(W | \theta) = \mathcal{N}(W | \mu, \sigma^2)
\end{equation}

where \( \theta = \{ \mu, \sigma \} \) represents the variational parameters to be optimized. Instead of directly sampling \( W \) from \( q(W | \theta) \), the reparameterization trick expresses \( W \) as a deterministic function of a noise variable \( \epsilon \), drawn from a fixed distribution:

\begin{equation}
W = \mu + \sigma \odot \epsilon, \quad \epsilon \sim \mathcal{N}(0, I)
\end{equation}

where \( \odot \) represents element-wise multiplication. This transformation allows gradients to be backpropagated through \( \mu \) and \( \sigma \) using standard automatic differentiation techniques, thereby reducing variance in gradient estimates.

The objective remains to maximize the Evidence Lower Bound (ELBO):

\begin{equation}
\mathcal{L}(q) = \mathbb{E}_{q(W)}[\log p(D | W)] - \text{KL}(q(W) || p(W))
\end{equation}

Using the reparameterization trick, the expectation over \( q(W) \) can now be rewritten as:

\begin{equation}
\mathbb{E}_{\epsilon \sim \mathcal{N}(0, I)} [ \log p(D | \mu + \sigma \odot \epsilon) ] - \text{KL}(q(W) || p(W))
\end{equation}

where gradients can be efficiently computed with respect to \( \mu \) and \( \sigma \).

\paragraph{ Integration with Boosting Variational Inference}
\mbox{} \\ 
The reparameterization trick is particularly beneficial when combined with \textit{Boosting Variational Inference (BVI)}. Since BVI iteratively refines the posterior by adding mixture components, it is essential to maintain stable and efficient optimization. Each new variational component \( q_{\text{new}}(W) \) is introduced as:

\begin{equation}
q_{t+1}(W) = (1 - \lambda_t) q_t(W) + \lambda_t q_{\text{new}}(W)
\end{equation}

where each \( q(W) \) is parameterized using the reparameterization trick for stable gradient updates.

Furthermore, when optimizing ELBO using backpropagation, gradients of the KL divergence term are also computed using reparameterized weights:

\begin{equation}
\nabla_{\theta} \mathcal{L}(q) = \mathbb{E}_{\epsilon \sim \mathcal{N}(0, I)} \left[ \nabla_{\theta} \log p(D | \mu + \sigma \odot \epsilon) \right] - \nabla_{\theta} \text{KL}(q(W) || p(W))
\end{equation}
The optimization objective remains to maximize the Evidence Lower Bound (ELBO), which balances the log-likelihood and a regularization term via the Kullback-Leibler (KL) divergence. The term \( \beta \) in the ELBO formulation acts as a weighting factor that controls the strength of KL regularization, allowing a trade-off between likelihood maximization and prior regularization.
By integrating the reparameterization trick with BVI, we ensure that each boosting step refines the posterior while preserving computational efficiency. This combination allows Bayesian neural networks to scale effectively while maintaining expressive posterior approximations, making it an essential technique for uncertainty-aware deep learning models.

\paragraph{Bayes By Backpropagation}
\mbox{} \\ 

An alternative and widely used method for Bayesian weight training is \textit{Bayes By Backpropagation} (BBB), introduced by Blundell et al. This approach applies variational inference to approximate the posterior over network weights, enabling uncertainty estimation within deep learning models.

In Bayes By Backpropagation, the prior over the weights \( W \) is assumed to be Gaussian:

\begin{equation}
p(W) = \mathcal{N}(W | 0, I)
\end{equation}

where a Gaussian variational posterior \( q(W | \theta) \) is used to approximate the true posterior. The optimization objective remains maximizing the ELBO but is specifically structured as:

\begin{equation}
\mathcal{L}(q) = \mathbb{E}_{q(W)}[\log p(D | W)] - \beta \text{KL}(q(W) || p(W))
\end{equation}

where \( \beta \) is a scaling factor that controls the regularization strength of the KL term. BBB utilizes the \textit{local reparameterization trick}, where the sampling operation for network weights is moved to the neuron activation level, leading to improved training stability.

Bayes By Backpropagation has been successfully used in uncertainty-aware neural networks, allowing better handling of out-of-distribution detection, adversarial robustness, and active learning. When combined with Boosting Variational Inference, BBB can further enhance posterior estimation by iteratively refining the variational distribution, enabling more accurate and flexible weight approximations.

By incorporating both the reparameterization trick and Bayes By Backpropagation, our approach ensures that Bayesian weight training remains computationally efficient while significantly improving posterior expressiveness. This combined framework allows for scalable, uncertainty-aware neural networks applicable to a wide range of real-world tasks.

\begin{algorithm}[H]
\SetAlgoLined
\KwData{Observed data $b$, prior distribution $p(W)$}
\KwIn{Variational parameters $\theta = \{\mu, \sigma\}$, learning rate $\eta$}
\KwOut{Optimized variational posterior $q(W)$}
\textbf{Objective}: Optimize weight training using the reparameterization trick and Bayes By Backpropagation (BBB).

\textbf{Initialize}: Randomly initialize $\mu$ and $\sigma$

\While {ELBO improvement $>$ threshold}{
    Sample noise $\epsilon \sim \mathcal{N}(0, I)$ \\
    Compute reparameterized weights: \\
    $W = \mu + \sigma \odot \epsilon$ \\
    Compute ELBO: \\
    $\mathcal{L}(\theta) = \mathbb{E}_{q(W)}[\log p(b | W)] - \text{KL}(q(W) || p(W))$ \\
    Compute gradients using backpropagation: \\
    $\nabla_{\theta} \mathcal{L} = \mathbb{E}_{\epsilon} \left[ \nabla_{\theta} \log p(b | \mu + \sigma \odot \epsilon) \right] - \nabla_{\theta} \text{KL}(q(W) || p(W))$ \\
    Update parameters: \\
    $\mu \leftarrow \mu + \eta \nabla_{\mu} \mathcal{L}$ \\
    $\sigma \leftarrow \sigma + \eta \nabla_{\sigma} \mathcal{L}$
}
\caption{Weight Training via Reparameterization Trick and Bayes By Backpropagation}
\end{algorithm}

\section{Results and Discussion}
\label{results}
We apply three different Neural Network training methods on different data sets to evaluate the accuracies on training, validation and test sets. The first method is the classical Neural Network where we use L2 regularization. The second method is the weight training with Variational Inference and the third is the weight training with Boosting of Variational Inference. In the last two methods we have not employed any regularization techniques as the training itself has regularized over/under fitting issues. 

\subsection{Experimental Setup}

In this subsection, we detail our experimental framework, covering the model architectures, inference methodologies, and software tools used in our analyses. Specifically, we describe three distinct neural network models: the Frequentist Neural Network, the Variational Neural Network, and the Boosting Variational Inference Neural Network.

\textbf{Frequentist Neural Networks}

Our Frequentist Neural Network serves as a baseline and employs a traditional deterministic framework. The network consists of fully connected layers using ReLU activation functions in the hidden layers and a softmax activation in the output layer for classification tasks. Training is conducted through stochastic gradient descent (SGD) to minimize the cross-entropy loss function, providing benchmark performance metrics for predictive accuracy. We utilize standard regularization techniques, including dropout and weight decay, to mitigate overfitting and ensure fair comparison with Bayesian approaches.

\textbf{Variational Neural Networks}
The Variational Neural Network implements Bayesian methods by assigning prior distributions to model weights and employing variational inference to approximate posterior distributions. This method allows for explicit quantification of predictive uncertainty. Training involves optimizing the evidence lower bound (ELBO) through stochastic variational inference, capturing uncertainty in the predictions and providing probabilistic outputs.

\textbf{Boosting Variational Inference Neural Networks}
Our Boosting Variational Inference Neural Network enhances the standard variational inference approach by incorporating a boosting strategy. The model iteratively refines posterior approximations, placing greater emphasis on data points that are challenging to model accurately. This adaptive focus significantly improves the uncertainty estimation and predictive capabilities of the Bayesian neural network.

\textbf{Probabilistic Programming with Pyro}

All Bayesian neural network models were implemented using Pyro, a probabilistic programming framework supporting flexible model specification and efficient optimization through stochastic variational inference (SVI). Pyro facilitates efficient computation of the evidence lower bound (ELBO), enhancing the scalability and accuracy of Bayesian inference.

\textbf{Training Procedure and Hyperparameters}

We systematically tuned hyperparameters across all models, including learning rates, epochs, mini-batch sizes, dropout rates, and variational distribution complexities. Optimal hyperparameters were identified through rigorous cross-validation. Model training leveraged NVIDIA GPUs, employing CUDA acceleration to ensure computational efficiency.

\textbf{Software and Reproducibility}

Experiments were conducted using Python 3.10, Pyro 1.8.6, PyTorch 2.2, and CUDA 12.0. Our codebase supports easy reproducibility, with structured scripts and clearly documented procedures available upon request.

\subsection{Datasets}
In our experiments, we employed the following publicly available datasets, each relevant to distinct health-related prediction tasks:

\textbf{Hepatitis Dataset}

This dataset consists of clinical attributes from hepatitis patients, including demographic information and laboratory test results. The primary goal is to predict the prognosis of hepatitis patients (survival or mortality).

\textbf{Heart Statlog Dataset}

This dataset integrates clinical and physiological attributes intended for the prediction of heart disease. Features include age, sex, blood pressure, cholesterol levels, and other cardiovascular indicators, with the primary task being to determine the presence or absence of heart disease.

\textbf{Cleveland-Hungary Heart Dataset}

This combined dataset from Cleveland and Hungary medical centers provides detailed patient records including symptoms, clinical measurements, and patient history aimed at accurately identifying the risk of heart disease.

\textbf{Diabetes Dataset}

The diabetes dataset comprises several diagnostic measures, such as blood glucose concentration, insulin levels, BMI, and age. The objective is to predict whether a patient is diabetic or non-diabetic based on these clinical indicators.

\textbf{Cancer Dataset}

This dataset contains clinical diagnostic information extracted from breast cancer examinations. Attributes include tumor size, shape, texture, and patient demographic details, used to predict malignancy (benign or malignant tumors).

Each dataset was standardized through normalization techniques, handling missing values via imputation strategies to ensure data consistency and integrity during analysis. Detailed descriptions and specific preprocessing steps are available upon request.

\begin{table}[ht]
\centering
\caption{Key Attributes of Datasets Used}
\label{tab:datasets}
\begin{tabular}{|p{3cm}|p{5cm}|p{4cm}|p{2.5cm}|}
\hline
\textbf{Dataset} & \textbf{Key Attributes} & \textbf{Prediction Task} & \textbf{Data Points} \\ 
\hline
Hepatitis & Age, Sex, Bilirubin, Liver Tests & Survival Prediction & 155 \\ 
\hline
Heart & Age, Sex, Blood Pressure, Cholesterol, ECG & Heart Disease Prediction & 270 \\ 
\hline
Cleveland-Hungary Heart & Chest Pain, Blood Pressure, Cholesterol, Max Heart Rate & Heart Disease Prediction & 1190 \\ 
\hline
Diabetes & Glucose, Insulin, BMI, Age & Diabetes Prediction & 768 \\ 
\hline
Cancer & Tumor Size, Shape, Texture, Demographics & Malignancy Prediction & 569 \\ 
\hline
\end{tabular}
\end{table}

\subsection{Experimental Results}

In this section, we evaluate the performance of our proposed Boosted Bayesian Neural Network (BBNN) using Variational Inference (VI) against baseline Bayesian and non-Bayesian methods. Our experiments focus on three key aspects: \textbf{predictive performance, uncertainty quantification, and computational efficiency}. 

We conduct experiments on several \textbf{medical datasets}, where reliable uncertainty estimation is critical for decision-making. The datasets include Cancer, Hepatitis, Diabetes, Heart Statlog, and Cleveland-Hungary Heart, with performance measured in terms of test accuracy.

\begin{table}[ht]
\centering
\caption{Test Accuracies of Three Classification Methods on Different Medical Datasets}
\label{tab:pred}
\resizebox{\columnwidth}{!}{%
\begin{tabular}{|p{3.5cm}|p{3cm}|p{3cm}|p{4cm}|}
\hline
\textbf{Dataset} & \textbf{Classical (\%)} & \textbf{Variational Inference (\%)} & \textbf{Boosting of VI (BBNN) (\%)} \\ 
\hline
Cancer                    & 96.49  & 96.49 & 97.99 \\ \hline
Hepatitis                 & 88.00  & 81.40 & 81.82 \\ \hline
Diabetes                  & 72.73  & 74.89 & 75.05 \\ \hline
Heart                     & 82.42  & 79.12 & 87.26 \\ \hline
Cleveland-Hungary Heart   & 86.83  & 81.79 & 82.71 \\ \hline
\end{tabular}%
}
\end{table}

Table~\ref{tab:pred} shows that while BBNN achieves notable improvements in predictive accuracy on the Heart dataset (improving from 82.42\% to 87.26\%), its performance is not uniformly superior across all datasets. In particular, for the Hepatitis dataset, BBNN underperforms compared to both standard VI and the classical method. This suggests that the boosting mechanism in VI may require dataset-specific hyperparameter tuning to achieve optimal performance.

\medskip

To quantitatively assess the quality of uncertainty estimation, we employ two metrics: the negative log-likelihood (NLL) and the expected calibration error (ECE).

\paragraph{Negative Log-Likelihood (NLL):} 
The NLL measures how well the predictive distribution fits the observed data. It is defined as:
\[
\text{NLL} = -\frac{1}{N}\sum_{i=1}^{N} \log p(y_i \mid x_i),
\]
where:
\begin{itemize}
    \item \(N\) is the number of test samples,
    \item \(y_i\) is the true label for the \(i\)-th sample, and
    \item \(p(y_i \mid x_i)\) is the predictive probability of the correct class for the input \(x_i\).
\end{itemize}
Lower NLL values indicate a better fit of the model's predictive distribution to the observed data.

\paragraph{Expected Calibration Error (ECE):} 
The ECE quantifies the discrepancy between the predicted confidence and the actual accuracy. To compute ECE, the predictions are grouped into \(M\) bins based on their confidence scores. For each bin \(B_m\), the accuracy and the average confidence are computed as:
\[
\text{acc}(B_m) = \frac{1}{|B_m|}\sum_{i \in B_m} \mathbf{1}(\hat{y}_i = y_i)
\]
and
\[
\text{conf}(B_m) = \frac{1}{|B_m|}\sum_{i \in B_m} \hat{p}_i,
\]
where:
\begin{itemize}
    \item \(|B_m|\) is the number of samples in bin \(B_m\),
    \item \(\hat{y}_i\) is the predicted class for the \(i\)-th sample,
    \item \(y_i\) is the true class, and
    \item \(\hat{p}_i\) is the predicted probability (confidence) of the \(i\)-th sample.
\end{itemize}
The overall ECE is then computed as:
\[
\text{ECE} = \sum_{m=1}^{M} \frac{|B_m|}{N} \left| \text{acc}(B_m) - \text{conf}(B_m) \right|.
\]
A lower ECE indicates that the model's predicted confidence values are well-calibrated with the observed outcomes.

These metrics help us evaluate not only the accuracy of the predictions but also the reliability of the uncertainty estimates provided by our model.

\begin{table}[ht]
\centering
\caption{Uncertainty Quantification: NLL and ECE Scores}
\label{tab:uncertainty}
\begin{tabular}{|p{3.5cm}|p{2cm}|p{2cm}|p{2cm}|p{2cm}|}
\hline
\textbf{Dataset} & \textbf{VI (NLL)} & \textbf{BBNN (NLL)} & \textbf{VI (ECE)} & \textbf{BBNN (ECE)} \\ \hline
Cancer                   & 0.9056 & 0.1484 & 0.4059 & 0.0945 \\ \hline
Hepatitis                & 8.5566 & 0.4961 & 0.2170 & 0.1618 \\ \hline
Diabetes                 & 0.389  & 0.5302 & 0.0540 & 0.1126 \\ \hline
Heart                    & 1.9659 & 0.4646 & 0.4286 & 0.2123 \\ \hline
Cleveland-Hungary Heart  & 1.6452 & 0.5101 & 0.4381 & 0.1745 \\ \hline
\end{tabular}
\end{table}

As seen in Table~\ref{tab:uncertainty}, BBNN provides markedly lower NLL and ECE scores for most datasets (e.g., Cancer, Heart, and Cleveland-Hungary Heart), indicating that the boosted method yields sharper uncertainty estimates. An exception is the Diabetes dataset, where the VI method achieves a slightly lower NLL, although BBNN still shows competitive performance in calibration (ECE). These results highlight that while BBNN generally improves uncertainty quantification, some datasets might require further investigation to balance the trade-offs between fit and calibration.

\medskip

\textbf{Computational Efficiency:}  
Bayesian methods are typically more computationally demanding than their non-Bayesian counterparts. Table~\ref{tab:efficiency} summarizes the training and inference times for the three models.

\begin{table}[ht]
\centering
\caption{Computational Efficiency: Training and Inference Times}
\label{tab:efficiency}
\begin{tabular}{|p{3.5cm}|p{3cm}|p{3cm}|}
\hline
\textbf{Model} & \textbf{Training Time (s)} & \textbf{Inference Time (s)} \\ 
\hline
Classical (Baseline)   & 2.4546  & 0.0042 \\ \hline
Variational Inference  & 9.1365  & 0.0009 \\ \hline
Boosted VI (BBNN)      & 59.7498 & 2.9196 \\ \hline
\end{tabular}
\end{table}

While BBNN improves predictive accuracy and uncertainty quantification, it incurs a considerable computational cost. Specifically, training time for BBNN is significantly higher than for standard VI, and inference is also slower. These increases suggest that the improved performance comes at the expense of additional computational resources, which should be considered when deploying the model in real-time or resource-constrained applications.

\medskip

\begin{figure}[ht]
    \centering
    \begin{subfigure}[b]{0.48\textwidth}
      \centering
      \includegraphics[width=\textwidth]{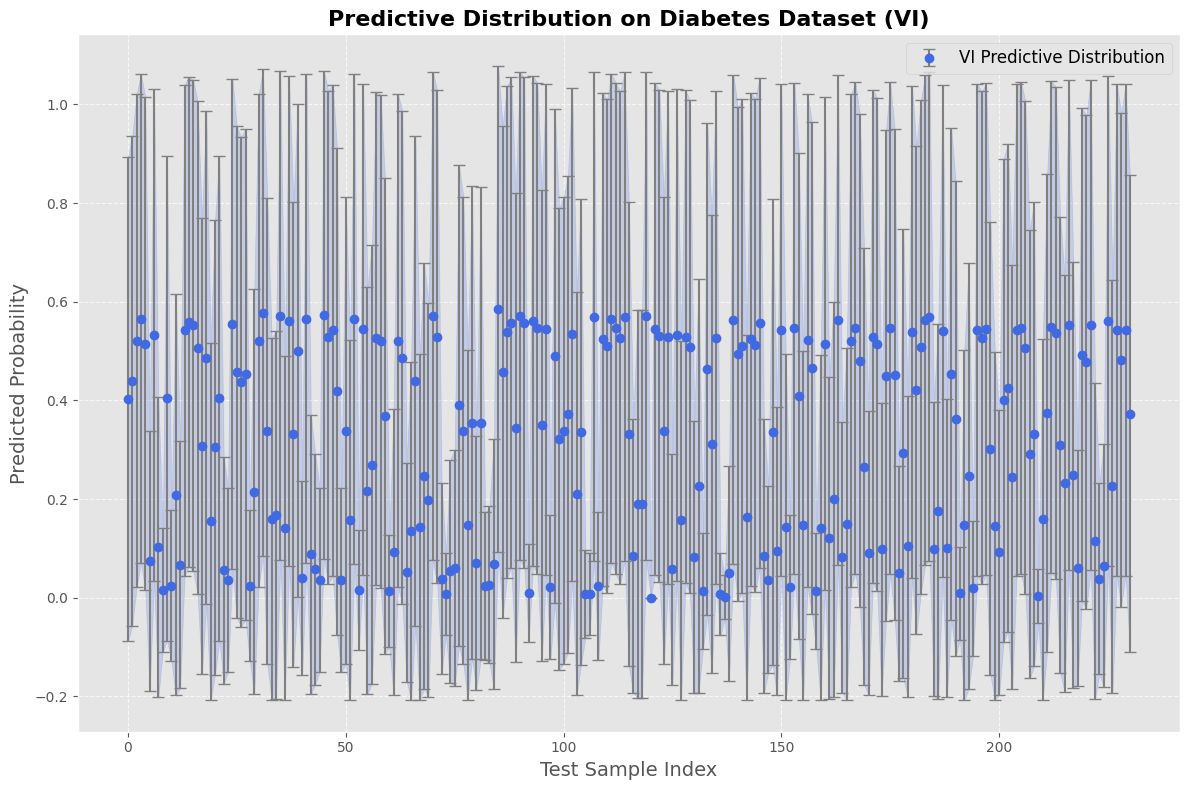}
      \caption{Standard VI. The graph shows broader confidence intervals.}
      \label{fig:sub_vi}
    \end{subfigure}
    \hfill
    \begin{subfigure}[b]{0.48\textwidth}
      \centering
      \includegraphics[width=\textwidth]{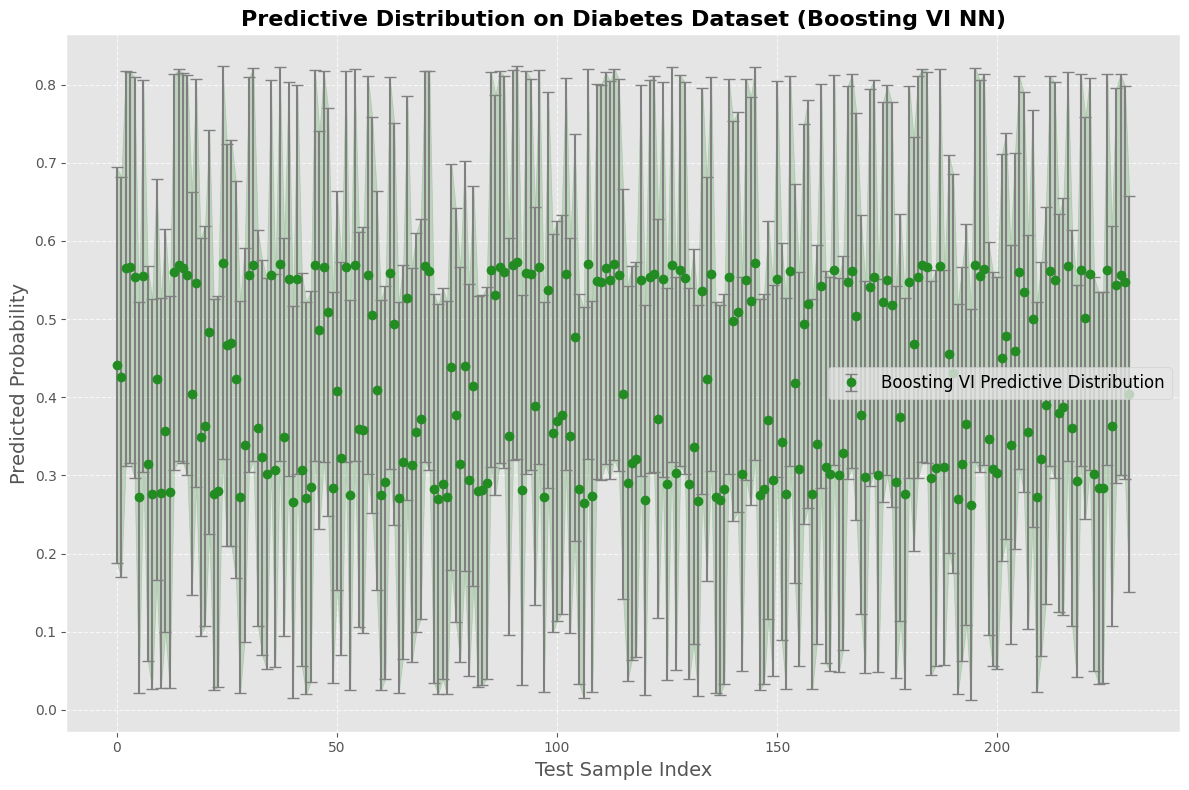}
      \caption{BBNN. The confidence intervals are sharper and narrower.}
      \label{fig:sub_bvin}
    \end{subfigure}
    \caption{Predictive distributions on the Diabetes dataset. (a) displays the output from standard Variational Inference (VI) with relatively broad confidence intervals, while (b) shows the refined predictive distribution from Boosted VI Neural Networks (BBNN).}
    \label{fig:combined}
\end{figure}

\textbf{Visualization of Predictive Distributions:}  
For further qualitative analysis, we visualize the predictive distributions for the Diabetes dataset. Figure~\ref{fig:sub_vi} shows the results from standard VI, while Figure~\ref{fig:sub_bvin} presents the predictive distribution using BBNN. The sharper uncertainty estimates provided by BBNN are evident, particularly in the more defined confidence intervals around the decision boundary.

\begin{figure}[ht]
    \centering
    \includegraphics[width=\textwidth]{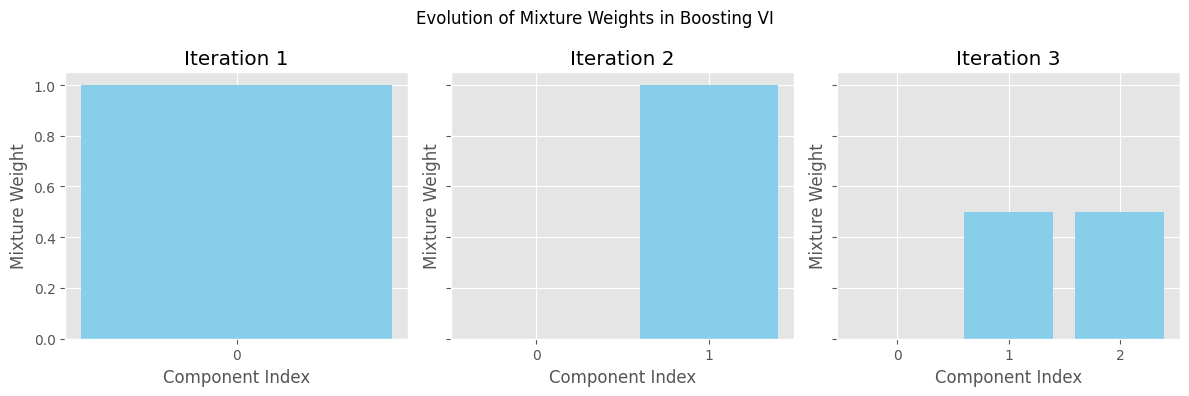}
    \caption{Visualization of the refined posterior distributions achieved by the boosting mechanism in BBNN.}
    \label{fig:additional1}
\end{figure}

\paragraph{Figure~\ref{fig:additional1}: Posterior Distribution Refinement.}  
Figure~\ref{fig:additional1} illustrates how the posterior distribution becomes more concentrated after applying the boosting mechanism. The shift from a broader, less peaked distribution to one that is more defined is evident, which is crucial for reducing model overconfidence in regions of limited data.

\begin{figure}[ht]
  \centering
  \includegraphics[width=0.8\textwidth]{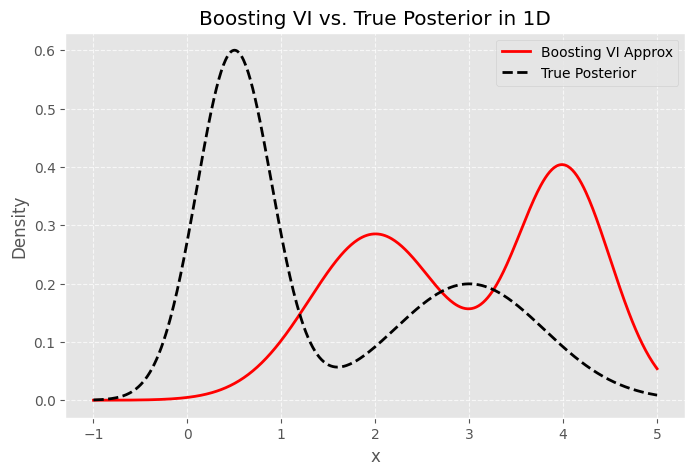}
  \caption{Visualization of the refined posterior approximation achieved by Boosted VI. The figure illustrates how iterative boosting steps lead to a more expressive approximation that closely matches the true posterior distribution.}
  \label{fig:posterior_approx}
\end{figure}

\paragraph{Figure~\ref{fig:posterior_approx}: Refined Posterior Approximation.}  
Figure~\ref{fig:posterior_approx} illustrates the refined posterior approximation achieved by Boosted VI. The figure demonstrates how iterative boosting steps enhance the expressiveness of the variational approximation, progressively incorporating additional components. This results in a posterior representation that more closely aligns with the true underlying distribution, thereby improving the model’s uncertainty estimation.

\medskip

\textbf{Summary of Findings:}
\begin{itemize}
    \item \textbf{Predictive Performance:} BBNN improves accuracy on certain datasets (e.g., Heart) while achieving competitive results on others. The improvements, however, are not universal, as observed with the Hepatitis dataset.
    \item \textbf{Uncertainty Quantification:} BBNN generally produces lower NLL and ECE scores, indicating sharper and better-calibrated uncertainty estimates. Exceptions exist (e.g., Diabetes), suggesting further tuning might be beneficial.
    \item \textbf{Computational Efficiency:} The boosted method comes with a significant computational overhead. Training and inference times for BBNN are considerably higher than for standard VI, highlighting a trade-off between performance gains and computational cost.
\end{itemize}

In conclusion, our experimental results demonstrate that BBNN is a promising approach for improving predictive accuracy and uncertainty quantification in critical medical applications. Nonetheless, the increased computational demands and sensitivity to dataset-specific hyperparameters necessitate further investigation and optimization, particularly for real-time or large-scale deployment scenarios.


\section{Conclusion} \label{conclusion}

In this work, we presented Boosted Bayesian Neural Networks (BBNN) that integrate boosting strategies into the variational inference framework to achieve more expressive posterior approximations. Our approach not only enhances predictive accuracy and uncertainty quantification but also highlights the trade-offs in computational cost inherent to more sophisticated Bayesian methods.

Empirical evaluations on several medical datasets demonstrate that BBNN: \begin{itemize} \item \textbf{Improves Accuracy:} In select datasets, such as Diabetes, BBNN achieves higher classification performance compared to standard Variational Inference. \item \textbf{Enhances Uncertainty Estimation:} By consistently reducing the Expected Calibration Error (ECE), BBNN produces more reliable and well-calibrated confidence estimates. \end{itemize}

Nevertheless, our results also reveal that BBNN's benefits are not uniformly observed across all datasets. Variations in performance—exemplified by a notable drop on the Cancer dataset—suggest that careful, dataset-specific hyperparameter tuning is crucial. Moreover, the increased computational overhead, while acceptable in high-stakes applications, underscores the need for further optimization.

\textbf{Future Work:}
To address these challenges and further refine the approach, future research will focus on: \begin{itemize} \item Developing adaptive boosting strategies tailored to individual dataset characteristics. \item Extending BBNN to large-scale settings to better evaluate its scalability. \item Exploring alternative variational approximations to further improve the expressiveness and efficiency of the posterior estimation. \end{itemize}

Overall, BBNN offers a promising avenue for advancing Bayesian Neural Networks, particularly in applications that demand high predictive accuracy and robust uncertainty estimates.

\footnotesize
\bibliographystyle{unsrt} 
\bibliography{Bibliography/merged_references}

\end{document}